\def\paperID{*****} 
\def\confName{CVPR}
\def\confYear{2026}

\def\paperTitle{Action–Geometry Prediction with 3D Geometric Prior for Bimanual Manipulation}

\def\authorBlock{
    Chongyang Xu$^{1,3}$\thanks{ This work was done during Chongyang Xu's internship at Dexmal.} \qquad
    Haipeng Li$^{2,3}$\thanks{Project Leader.} \qquad
    Shen Cheng$^{3}$ \qquad 
    Jingyu Hu$^{4}$ \qquad \\
    Haoqiang Fan$^{3}$ \qquad
    Ziliang Feng$^{1}$ \qquad 
    Shuaicheng
    Liu$^{2}$\thanks{Corresponding Author: liushuaicheng@uestc.edu.cn.} \\
    $^{1}$ College of Computer Science, Sichuan University, China \\
    $^{2}$ University of Electronic Science and Technology of China, China \\
    $^{3}$ Dexmal \hspace{0.5em} $^{4}$The Chinese University of Hong Kong, HK SAR, China
}

\newif\ifreview 
\newif\ifarxiv \newcommand{\arxiv}{\arxivtrue}
\newif\ifcamera 
\newif\ifrebuttal 

\arxiv 

\pdfoutput=1
\documentclass[10pt,twocolumn,letterpaper]{article}
\ifreview \usepackage[review]{cvpr} \fi
\ifarxiv \usepackage[pagenumbers]{cvpr} \fi
\ifrebuttal \usepackage[rebuttal]{cvpr} \fi
\ifcamera \usepackage{cvpr} \fi

\usepackage{graphicx}	
\usepackage{amsmath}	
\usepackage{amssymb}	
\usepackage{booktabs}
\usepackage{times}
\usepackage{microtype}
\usepackage{epsfig}
\usepackage{caption}
\usepackage{float}
\usepackage{placeins}
\usepackage{color, colortbl}
\usepackage{stfloats}
\usepackage{enumitem}
\usepackage{tabularx}
\usepackage{xstring}
\usepackage{multirow}
\usepackage{xspace}
\usepackage{url}
\usepackage{subcaption}
\usepackage{xcolor}
\usepackage[hang,flushmargin]{footmisc}

\usepackage{tikz}
\usepackage{pgfplots}
\pgfplotsset{compat=1.17}

\usepackage{booktabs}
\usepackage{multirow}
\usepackage{graphicx}
\usepackage{makecell}
\usepackage{algorithm}
\usepackage{algorithmic}
\usepackage{amsmath} 
\usepackage{amssymb}
\usepackage{pifont}
\usepackage{xcolor}

\ifcamera \usepackage[accsupp]{axessibility} \fi

% ...

\ifarxiv  \fi

\newcommand{\R}[1]{{%
    \textbf{%
        \ifstrequal{#1}{1}{\textcolor{red}{R#1}}{%
        \ifstrequal{#1}{2}{\textcolor{blue}{R#1}}{%
        \ifstrequal{#1}{3}{\textcolor{magenta}{R#1}}{%
        \ifstrequal{#1}{4}{\textcolor{teal}{R#1}}{%
                           \textcolor{cyan}{R#1}%
        }}}}%
    }%
}}

\usepackage{xr-hyper}

\makeatletter
\newcommand*{\addFileDependency}[1]{
  \typeout{(#1)}
  \@addtofilelist{#1}
  \IfFileExists{#1}{}{\typeout{No file #1.}}
}

\makeatother
\newcommand*{\myexternaldocument}[1]{
    \externaldocument{#1}
    \addFileDependency{#1.tex}
    \addFileDependency{#1.aux}
}

\definecolor{cvprblue}{rgb}{0.21,0.49,0.74}
\usepackage[pagebackref,breaklinks,colorlinks,allcolors=cvprblue]{hyperref}
\usepackage[capitalize]{cleveref}
\crefname{section}{Sec.}{Secs.}
\crefname{table}{Table}{Tables}
\crefname{figure}{Fig.}{Figs.}

\ifarxiv \crefname{appendix}{App.}{Apps.}
\else \crefname{appendix}{Suppl.}{Suppls.} \fi

\unless\ifarxiv \myexternaldocument{_supplementary} \fi

\begin{document}

\title{\paperTitle}
\author{\authorBlock}
\maketitle

\begin{abstract}
Bimanual manipulation requires policies that can reason about 3D geometry, anticipate how it evolves under action, and generate smooth, coordinated motions. 
However, existing methods typically rely on 2D features with limited spatial awareness, or require explicit point clouds that are difficult to obtain reliably in real-world settings. 
At the same time, recent 3D geometric foundation models show that accurate and diverse 3D structure can be reconstructed directly from RGB images in a fast and robust manner.
We leverage this opportunity and propose a framework that builds bimanual manipulation directly on a pre-trained 3D geometric foundation model. 
Our policy fuses geometry-aware latents, 2D semantic features, and proprioception into a unified state representation, and uses diffusion model to jointly predict a future action chunk and a future 3D latent that decodes into a dense pointmap. 
By explicitly predicting how the 3D scene will evolve together with the action sequence, the policy gains strong spatial understanding and predictive capability using only RGB observations.
We evaluate our method both in simulation on the RoboTwin benchmark and in real-world robot executions. 
Our approach consistently outperforms 2D-based and point-cloud-based baselines, achieving state-of-the-art performance in manipulation success, inter-arm coordination, and 3D spatial prediction accuracy. Code is available at \url{https://github.com/Chongyang-99/GAP.git}.
\end{abstract}
\section{Introduction}
\label{sec:intro}

\begin{figure}[t]
    \centering
    \includegraphics[width=\linewidth]{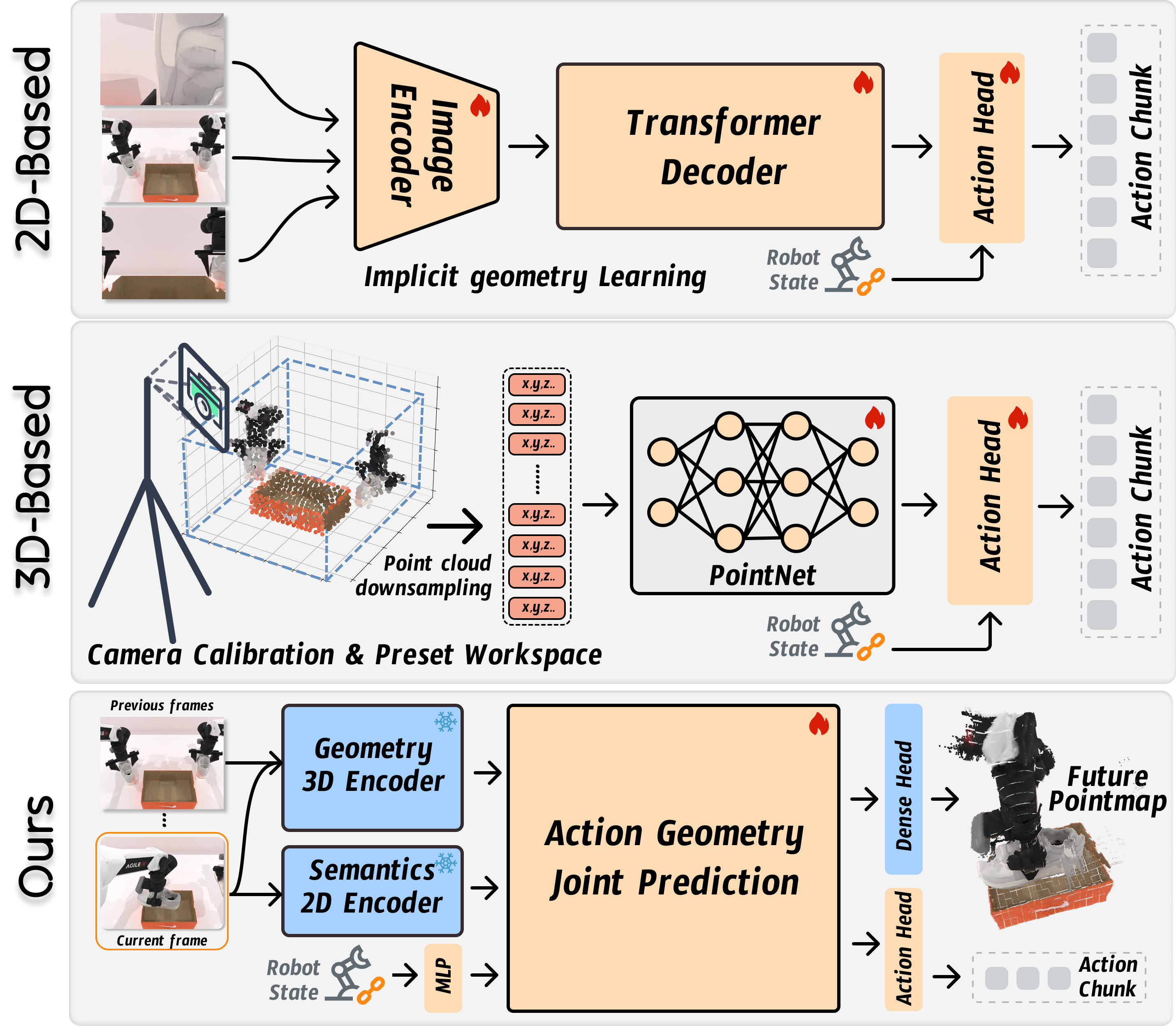}
    \caption{\textbf{Paradigm Comparison.} 2D-based methods learn implicit 3D representations from multi-view RGB observations, relying purely on 2D cues. 3D-based methods require camera calibration and preset workspaces to crop point clouds, which limits generalization and scalability. In contrast, our approach leverages powerful 2D and 3D pretrained priors to achieve semantic–geometric fusion perception, enabling robust action and geometry joint prediction without strict calibration or workspace constraints.}
    \label{fig:teaser}
\end{figure}

Bimanual manipulation~\cite{zhao2023learning, mu2025robotwin} equips robots with the ability to perform coordinated motions that exceed the capabilities of single-arm systems. This capability is crucial for precision assembly~\cite{Choi2016ProbabilisticVV, Yamazaki2018AssemblyMU, Collins2023RAMPAB}, deformable-object handling~\cite{Bao2023DexArtBG, zheng2025xvlasoftpromptedtransformerscalable}, and operation in cluttered or dynamic environments~\cite{black2024pi0visionlanguageactionflowmodel, intelligence2025pi05visionlanguageactionmodelopenworld}. 
However, reliable bimanual manipulation remains challenging, requiring the policy to generate temporally smooth, dynamically consistent actions~\cite{zhao2023learning, black2024pi0visionlanguageactionflowmodel}, perceive 3D object geometry and interaction dynamics~\cite{ze20243d, chen2025g3flow, liang2025dexhanddiff, weng2024dexdiffuser}, and maintain stable inter-arm coordination throughout continuous contact~\cite{lee2024interact, jiang2025rethinkingbimanualroboticmanipulation}.

Recent policy-learning approaches such as ACT~\cite{zhao2023learning} and diffusion policies~\cite{chi2023diffusion} improve stability by combining action chunking with iterative denoising. While effective for temporal smoothing, these methods remain largely \emph{geometry-flat}: they rely on 2D features or low-level observations without explicit 3D structure (as illustrated in the top of Fig.~\ref{fig:teaser}), limiting their ability to reason about spatial relationships, occlusions, and contact-rich interactions.

To introduce 3D awareness, DP3~\cite{ze20243d} demonstrates that conditioning policies on point clouds yields more faithful geometric reasoning (as illustrated in the middle of Fig.~\ref{fig:teaser}). Subsequent works~\cite{wang2023d, wang2024gendp, jia_lift3d_2024, chen2025g3flow} build on this insight by incorporating 3D cues together with semantic priors; for example, G3Flow~\cite{chen2025g3flow} lifts 2D features into 3D for improved semantic–geometric alignment. However, these pipelines rely on supervision or sensor inputs that are difficult to obtain reliably in real-world settings. High-quality point clouds are straightforward to access in simulation but require careful calibration and are sensitive to noise and occlusion. Meanwhile, 2D-to-3D lifting approaches~\cite{jia_lift3d_2024, wang2024gendp} depend on sparse back-projection or optimization, often yielding low-resolution structure and substantial engineering overhead.

Recently, 3D geometric models~\cite{wang2025vggt,dust3r_cvpr24,wang2025pi3} have shown that robust 3D assets can be reconstructed {directly from RGB images}, without depth sensors nor explicit point clouds. Given a set of images, these models produce dense pointmaps and other geometric outputs in a fast feed-forward manner. Motivated by these advances, recent work increasingly explores injecting stronger 3D cues into manipulation~\cite{ge2025vggtdpgeneralizablerobotcontrol, qian2025gp33dgeometryawarepolicy}. Yet, leveraging geometric foundation models as a high-capacity, feed-forward geometric prior, particularly one that directly supports predictive 3D reasoning for bimanual control, remains an emerging and promising direction. We explore this avenue by coupling action generation with future 3D structure prediction, using the geometric model as the core prior linking perception and control.

This motivates our central question: \emph{can a bimanual policy directly leverage a 3D geometric foundation model to achieve RGB-only, 3D-aware predictive control, without explicit point cloud pipelines?} 
To answer this, we propose an end-to-end framework that uses a pre-trained 3D foundation model as the perception backbone and jointly learns to predict both future actions and future 3D structure, demonstrated in the bottom of Fig.~\ref{fig:teaser}. 
Specifically, the multi-view model $\pi^3$~\cite{wang2025pi3} encodes temporal RGB observations into a geometry-aware latent representation, which is fused with 2D semantic features and proprioceptive state to form a unified context for control. 
A conditional diffusion policy then jointly denoises a future action chunk together with a 3D latent that decodes into a dense pointmap. 
By predicting the 3D scene state alongside the action sequence, the policy learns to anticipate how spatial relationships will evolve under the influence of those actions.

We evaluate our method both in simulation on the RoboTwin 2.0 benchmark~\cite{mu2025robotwin} and in real-world robot executions. Our experiments demonstrate that leveraging 3D foundation-model features and jointly predicting future 3D structure substantially improves manipulation success, inter-arm coordination, and geometric consistency, while avoiding explicit point cloud acquisition or calibration.

Our main contributions are summarized as follows:
\begin{itemize}
    \item We leverage a pre-trained 3D geometric foundation model as the core perception prior for bimanual manipulation, enabling RGB-only 3D-aware policy learning without explicit point clouds.

    \item We introduce explicit future 3D prediction by having the policy generate a 3D latent that decodes into a dense pointmap, allowing the policy to reason about how the scene geometry will evolve under its actions.

    \item Through extensive experiments in simulation and the real world, we show consistent improvements over 2D-based and point-cloud-based baselines, as our method's enhanced inter-arm coordination and 3D spatial perception directly contribute to higher manipulation success rates.
\end{itemize}

\section{Related Work}
\label{sec:related}

\subsection{Imitation Learning for Manipulation}



Effective robotic manipulation, particularly for bimanual, contact-rich operations, demands policies that generate smooth, temporally coherent trajectories and leverage precise geometric perception~\cite{Gao2024BiKVILKV, mu2025robotwin, zhao2023learning, liu_rdt-1b_2025, 10343126, lee2024interact, wang2025oneshot}. Early methods addressed trajectory smoothness using techniques like action-chunking, which proved effective at imitating expert distributions~\cite{zhao2023learning, lee2024interact, 10802845}. More recent diffusion-based policies advanced this by formulating trajectory generation as a sampling process, thereby capturing the intrinsic multimodality of manipulation tasks~\cite{chi2023diffusion, liang2023adaptdiffuser, liu_rdt-1b_2025, prasad2024consistency, zhang2025flowpolicy}.

Concurrently, input representations have progressively evolved from conventional 2D imagery to richer and more expressive 3D data. Models such as DP3~\cite{ze20243d} demonstrated the distinct advantages of point cloud inputs, and numerous subsequent works~\cite{jia_lift3d_2024, chen2025g3flow, wang2023d, wang2024gendp, zhu2025spa} have increasingly leveraged 3D cues to enhance crucial spatial understanding for improved manipulation success. However, a significant limitation of these 3D-centric approaches is their reliance on computationally preprocessing and meticulous calibration, which severely hinders widespread practical deployment.

\subsection{Pretrained Visual Representations}

Large-scale pre-trained visual backbones~\cite{oquab2023dinov2, simeoni2025dinov3, radford2021learning} provide expressive features widely adopted for semantic guidance in manipulation pipelines~\cite{chen2025g3flow, zhu2024densematcher, wang2024gendp}. 
Concurrently, large-scale pretraining of feed-forward 3D reconstruction models~\cite{dust3r_cvpr24, wang2025vggt} has yielded robust, geometry-aware representations that scale effectively to large 3D datasets. Subsequent extensions to multi-view~\cite{wang2025pi3} and temporal settings~\cite{chen2025easi3r, cut3r, chen2025ttt3r} further enhanced the spatio-temporal consistency of reconstructed scenes.

These powerful pretrained 3D backbones were then adopted directly in manipulation. Rich features from models like VGGT~\cite{wang2025vggt}, for instance, are now successfully used for action denoising and policy learning~\cite{qian2025gp33dgeometryawarepolicy, ge2025vggtdpgeneralizablerobotcontrol, vuong2025improvingroboticmanipulationefficient}, demonstrating effectively that pretrained geometric priors can benefit both control and perception. However, such purely geometry-centric representations primarily capture static structural cues and often lack crucial high-level semantics regarding task intent or complex inter-agent relations. This severely limits their effectiveness in highly complex cooperative or long-horizon sequential manipulation tasks.

\subsection{Predictive World Models}

World models learn compact representations of environmental dynamics, enabling agents to reason about future states and the consequences of their actions~\cite{Genie, agarwal2025cosmos}. These models offer significant advantages in robotic manipulation by providing structured priors for policy learning~\cite{FangqiIRASim2024, xu2025imagineact, zhang2024pivotr, hansen2024tdmpc2}.

Early approaches simply focused on predicting future observations directly in the conventional 2D image space~\cite{du2023learning, wu2024unleashing, black2024zeroshot}. More recent methods introduced compact latent world models~\cite{agarwal2025cosmos}, which first efficiently compress high-dimensional visual inputs and then predict future states within this highly compressed latent space~\cite{xu2025imagineact}. Despite notable variations in representation, these methods still operate fundamentally in the 2D observation space. This heavy reliance on 2D inputs, without crucial explicit 3D structural reasoning, often results in predictions that lack sufficient geometric consistency and physical plausibility, especially for tasks involving highly complex spatial interactions.

\section{Method}
\label{sec:method}
\begin{figure*}[t]
    \centering
    \includegraphics[width=\linewidth]{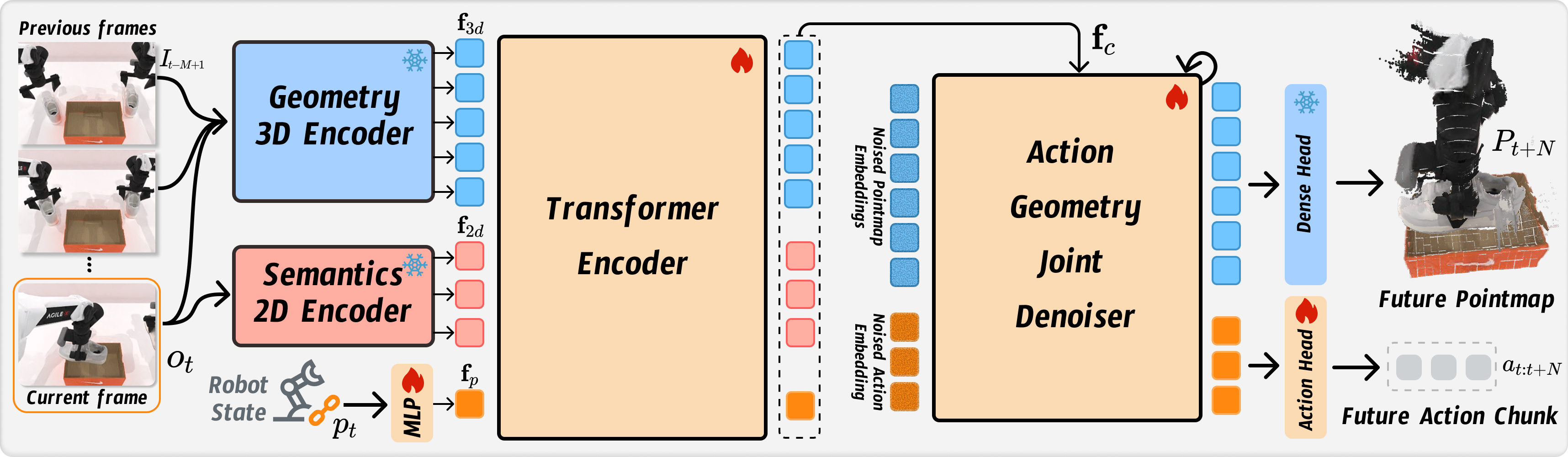}
    \caption{\textbf{Overview of our method.} Given a sequence of past RGB frames, the current image, and proprioceptive state, our model extracts 3D geometric features, 2D semantic features, and robot state embeddings through three parallel encoders. 
    These signals are fused by a Transformer into a unified semantic and geometric context that conditions a joint denoising process. 
    A conditional diffusion decoder then predicts both a future action chunk and a future 3D latent, which is further decoded into a dense pointmap.}

    \label{fig:pipeline}
\end{figure*}

\subsection{Preliminaries}

\textbf{Bimanual Learning.}
We aim to learn a bimanual policy $\pi$ from expert demonstrations, given as a dataset of trajectories
$\mathcal{D} = \{\tau_i\}$. 
Each trajectory $\tau_i$ is a sequence of environment observations and robotic actions:
\begin{equation}
    \tau_i = \{(o_1, p_1, a_1), \dots, (o_L, p_L, a_L)\},
\end{equation}
where $o_t$ is the visual observation, $p_t$ is the robot proprioceptive state at time $t$, and 
$a_t = (a_t^l, a_t^r)$ is the bimanual action containing the joint positions of the left and right arms, as well as their gripper states. 

Given the current state $(o_t, p_t)$, the policy predicts a sequence of $N$ future actions:
\begin{equation}
    \pi(o_t, p_t) \rightarrow a_{t:t+N},
\end{equation}
where $a_{t:t+N} = \{a_t, a_{t+1}, \dots, a_{t+N-1}\}$ denotes an action chunk of length $N$.

\noindent\textbf{3D Feed-forward Prediction.}
To provide rich 3D-aware features for our policy, we leverage $\pi^3$~\cite{wang2025pi3}. 
$\pi^3$ is a feed-forward network that takes a collection of $M$ RGB images $\{I_i\}_{i=1}^{M}$ observing the same 3D scene and processes them jointly with a Transformer backbone. 
The backbone outputs refined features that encode multi-view geometry, which are then passed into prediction heads to produce diverse 3D assets, including pointmaps, depth maps, etc. 

\subsection{Overview}

Our method, illustrated in Fig.~\ref{fig:pipeline}, is a multi-modal conditional generative model that jointly predicts future actions and 3D geometry. 
Given the current state, we first construct a unified semantic-geometric representation, which then conditions a joint denoising process to generate future action chunks and 3D latents.

At each time step $t$, the model takes three inputs:
(i) a sequence of previous RGB frames $V$, 
(ii) the current RGB frame $I_t$, and 
(iii) the current robot proprioceptive state $p_t$. 
These correspond to the state $(o_t, p_t)$, with $o_t = \{V, I_t\}$.
The inputs are processed by three parallel encoders:
\begin{itemize}
    \item \textbf{Geometry 3D Encoder.} Processes the visual observation $\{V, I_t\}$ to extract a 3D geometric feature $\mathbf{f}_{3d}$;
    \item \textbf{Semantics 2D Encoder.} Processes the current frame $I_t$ to extract a semantic feature $\mathbf{f}_{2d}$;
    \item \textbf{State Encoder.} An MLP encodes the proprioceptive state $p_t$ into a state embedding $\mathbf{f}_{p}$.
\end{itemize}

The three heterogeneous features $[\mathbf{f}_{3d}, \mathbf{f}_{2d}, \mathbf{f}_{p}]$ are concatenated along the token dimension and fed into a Transformer encoder, yielding a unified \emph{Semantic-Geometric Fused Context} $\mathbf{f}_{c}$. 
This context $\mathbf{f}_{c}$ conditions our generative module, the \emph{Joint Action--Geometry Denoiser}, which jointly denoises the latent representations for both actions and 3D geometry. 
Finally, task-specific heads produce the final predictions: 1) a {Dense Head} decodes the geometry latent into a pointmap, while 2) an {MLP Head} decodes the action latent into the action chunk $a_{t:t+N}$.

\subsection{Model Architecture}
\label{sec:baseline_arch}

Our pipeline is built upon ACT~\cite{zhao2023learning} and \emph{Xu et al.}~\cite{xu2025imagineact}. 
Unlike their train-from-scratch ResNet image encoder, we exploit large-scale pre-trained foundation models to separately extract 2D semantic and 3D geometric knowledge. 
We first describe the feature extraction and fusion stage.

\noindent\textbf{Geometry 3D Encoder.}
To obtain dynamic 3D geometric features $\mathbf{f}_{3d}$, we process the temporal visual observation $\{V, I_t\}$. 
We uniformly sample 5 frames from the past frames $V$ and concatenate them with the current frame $I_t$, forming a 6-frame sequence. 
This sequence is fed into a $\pi^3$~\cite{wang2025pi3} encoder. 
Following $\pi^3$, each frame is patchified into $14 \times 14$ patches. 
We extract features from the last two layers of the $\pi^3$ backbone (excluding decoding heads) and concatenate them to a 1024-dimensional geometric feature vector $\mathbf{f}_{3d}$.

\noindent\textbf{Semantics 2D Encoder.}
For 2D semantic features $\mathbf{f}_{2d}$, we process the current frame $I_t$ using a 2D foundation model, i.e., DINOv3~\cite{simeoni2025dinov3} encoder. The frame $I_t$ is divided into $16 \times 16$ patches and passed through the encoder, yielding a 1024-dimensional semantic feature vector $\mathbf{f}_{2d}$.

\noindent\textbf{State Encoder.}
The robot proprioceptive state $p_t \in \mathbb{R}^{14}$ encodes the bimanual configuration: two 7-dimensional vectors for the left and right arms. 
Each 7D vector contains 6 joint rotation angles and 1 gripper state. 
We project $p_t$ with a simple MLP into a 1024-dimensional state embedding $\mathbf{f}_{p}$.

\noindent\textbf{Semantic-Geometric Fusion.}
The three 1024-dimensional features $[\mathbf{f}_{3d}, \mathbf{f}_{2d}, \mathbf{f}_{p}]$ are concatenated into a token sequence and fed into a 4-layer DETR encoder~\cite{carion2020end} for deep fusion. 
The encoder output is the final Semantic-Geometric Fused Context $\mathbf{f}_{c}$, which serves as the conditions.

\noindent\textbf{Joint Diffusion Decoder.}
Given $\mathbf{f}_{c}$, we adopt a conditional diffusion decoder with an action chunking strategy following ACT~\cite{zhao2023learning}. 
The decoder backbone is implemented as a standard DETR decoder. 
During training, Gaussian noise is added to the ground-truth targets, and the decoder is trained to denoise the noisy inputs towards the clean targets. 
At inference time, the model starts from Gaussian noise and iteratively denoises it to generate actions and 3D latents, conditioned on $\mathbf{f}_{c}$.

\noindent\textbf{Heterogeneous Prediction Targets.}
The decoder jointly predicts two heterogeneous outputs:

\begin{itemize}
    \item \textbf{Future Action Chunk.}  
    This is the control output. 
    Following ACT~\cite{zhao2023learning}, we predict a chunk of $N$ future actions, forming a tensor in $\mathbb{R}^{N \times 14}$, where each 14D vector corresponds to a future proprioceptive state (6-DoF joints + 1-DoF gripper for each of the two arms).
 
    \item \textbf{Future 3D Pointmap.} To provide the model with explicit 3D awareness, we introduce a future 3D pointmap target, $P_{t+N}$. This pointmap is predicted at horizon $N$ by decoding an 3D latent embedding $\mathbf{f}_{t+N}$:
    \begin{equation}
        P_{t+N} = \mathrm{Dec}(\mathbf{f}_{t+N}),
    \end{equation}
    where $\mathbf{f}_{t+N} \in \mathbb{R}^{H/14 \times W/14 \times 1024}$ and $P_{t+N} \in \mathbb{R}^{H \times W \times 4}$. The first three channels of $P_{t+N}$ represent $(x,y,z)$ coordinates, while the fourth contains a per-point confidence.
\end{itemize}

\noindent We hypothesize that explicitly predicting future 3D geometry forces the model to acquire robust 3D-aware representations, which in turn promotes physically plausible and accurate long-horizon action plans. 
We empirically validate the importance of this joint 3D prediction in Sec.~\ref{sec:ablation}.

\subsection{Training}
\label{sec:training}

We follow the standard diffusion framework, but employ a joint supervision strategy. 
Instead of directly supervising the dense pointmap $P_{t+N}$, we jointly supervise both $P_{t+N}$ and its compact 3D latent embedding.
We denote the set of clean training targets as
\begin{equation}
    x_0 = \{ a_{t:t+N}, \mathbf{f}_{t+N}, P_{t+N} \},
\end{equation}
where $a_{t:t+N}$ is the future action chunk and $\mathbf{f}_{t+N}$ is the 3D latent corresponding to the future pointmap $P_{t+N}$.

\noindent\textbf{Forward Diffusion.}
The forward diffusion process $q$ gradually adds Gaussian noise to $x_0$. 
For a given diffusion step $k \in \{1, \dots, T\}$ (where $T$ is the total number of diffusion steps), a noisy sample $x_k$ can be written in closed form as
\begin{equation}
    q(x_k \mid x_0) = \mathcal{N}\!\left(x_k; \sqrt{\bar{\alpha}_k}\, x_0,\ (1 - \bar{\alpha}_k)\mathbf{I}\right),
    \label{eq:forward_diffusion}
\end{equation}
or equivalently,
\begin{equation}
    x_k = \sqrt{\bar{\alpha}_k}\, x_0 + \sqrt{1 - \bar{\alpha}_k}\, \epsilon,
    \label{eq:forward_sample}
\end{equation}
where $\epsilon \sim \mathcal{N}(0, \mathbf{I})$ and 
$\bar{\alpha}_k = \prod_{i=1}^{k} (1 - \beta_i)$ is the cumulative product of the noise schedule.

\noindent\textbf{Reverse Denoising.}
The Joint Action--Geometry Denoiser (our DETR decoder) takes as input the noisy targets $x_k$, the fused context $\mathbf{f}_{c}$, and the diffusion step $k$, and predicts the clean data $\hat{x}_0 =\{ \hat{a}_{t:t+N}, \hat{\mathbf{f}}_{t+N}, \hat{P}_{t+N} \}$. 
We optimize it with an L1 loss between $\hat{x}_0$ and $x_0$:
\begin{equation}
\begin{split}
    \mathcal{L} = &\mathbb{E}_{k, x_0, \epsilon} \Biggl[\left\| \hat{a}_{t:t+N} - a_{t:t+N} \right\|_1 + \\
    & \lambda \left\| \hat{\mathbf{f}}_{t+N} - \mathbf{f}_{t+N} \right\|_1 + \gamma \left\| \hat{P}_{t+N} - P_{t+N} \right\|_1 \Biggr],
\end{split}
\end{equation}
where $\lambda$ and $\gamma$ are balancing hyperparameters.

\noindent\textbf{Motivation for 3D Latent Prediction.}
We deliberately predict the compact 3D latent $\mathbf{f}_{t+N}$ and pointmap $P_{t+N}$ jointly. By supervising the model only on the final latent at horizon $N$, we force it to infer the complete 3D scene state resulting from the entire action sequence. This "look-ahead" mechanism is crucial for effective long-horizon planning.

\noindent\textbf{Pseudo-Ground Truth Generation.}
To obtain stable supervision for $\mathbf{f}_{t+N}$, we pre-extract 3D latents from all expert demonstrations using $\pi^3$. 
A naive approach that feeds a single frame $I_t$ into $\pi^3$ produces noisy and unstable features. 
Instead, we adopt a temporal observation window. 
For every frame index $s$ in the dataset, we uniformly sample $n$ preceding frames from its history and feed the sequence $\{V, I_s\}$ into the Geometry 3D Encoder ($\pi^3$). 
We then retain only the latent embedding $\mathbf{f}_s$ corresponding to the current frame $I_s$. 
For a training sample at time $t$ with horizon $N$, the 3D target is set to $\mathbf{f}_{t+N}$. 
This joint temporal processing substantially stabilizes the 3D latents and provides a clean pseudo ground-truth signal for training.

\subsection{Inference}
\label{sec:inference}

At inference time, predictions are generated by reversing the diffusion process. We first sample a pure Gaussian noise tensor $\tilde{x}_T \sim \mathcal{N}(0, \mathbf{I})$ and compute the Semantic-Geometric Fused Context $\mathbf{f}_{c}$ from the current inputs $\{V, I_t, p_t\}$ using the previously described encoders. The Joint Action--Geometry Denoiser then iteratively refines $\tilde{x}_T$ for $K$ denoising steps ($K \le T$) to produce the final clean prediction:
\begin{equation}
\hat{x}_0 = \{ \hat{a}_{t:t+N}, \hat{\mathbf{f}}_{t+N}, P_{t+N} \}.
\end{equation}
From this output, the action chunk $\hat{a}_{t:t+N}$ is executed by the robot. To improve inference efficiency, the decoding of $P_{t+N}$ can be optionally skipped.

\section{Experiments}
\label{sec:experiments}

We validate our method through extensive experiments in both simulation and the real world, and we aim to answer the following key research questions:
\begin{itemize}
    \item\textbf{RQ1:} Can a \textbf{pre-trained 3D backbone}, when operating on only 2D inputs, achieve 3D perception performance that is superior to \textbf{baseline methods}? (Sec. \ref{sec:sota_comp})
    
    \item\textbf{RQ2:} How important are the \textbf{future pointmap prediction paradigm} for 3D-aware perception? (Sec. \ref{sec:ablation})
    
    \item\textbf{RQ3:} Can our model \textbf{maintain its performance and generalize effectively} when deployed on real robots, beyond controlled simulation environments? (Sec. \ref{sec:realworld_comp})

\end{itemize}

\subsection{Baselines}


We evaluate our approach against several state-of-the-art visuomotor imitation learning baselines, representing a progression from 2D RGB inputs to 3D representations, to validate the advantages of our method across varying perceptual modalities and increasing geometric fidelity:

\begin{itemize}
    \item We first consider 2D-based policies. 
    \textbf{ACT}~\cite{zhao2023learning} employs a DETR-inspired Transformer to predict \textit{action chunks} from multi-view images, mitigating compounding errors. 
    \textbf{Diffusion Policy (DP)}~\cite{chi2023diffusion} formulates action generation as a conditional denoising diffusion process over single-view images, adeptly handling multi-modal action distributions.

    \item \textbf{3D Diffusion Policy (DP3)}~\cite{ze20243d} extends DP by operating directly on point cloud data via an efficient point encoder, capitalizing on 3D geometry. 
    \textbf{G3Flow}~\cite{mu2025robotwin} further enhances DP3 by explicitly integrating semantics, projecting foundation model features onto the point cloud to create a \textit{semantic flow} that maintains object-centric information.

    \item \textbf{RDT}~\cite{liu_rdt-1b_2025} is a 1.2B parameter diffusion-based Transformer, designed as a foundation model for bimanual manipulation using multi-view images. 
    \textbf{\textit{Xu et al.}}~\cite{xu2025imagineact} tackles bimanual coordination by jointly optimizing action prediction and future frames prediction.
\end{itemize}

\begin{figure}[t]
    \centering
    \includegraphics[width=\linewidth]{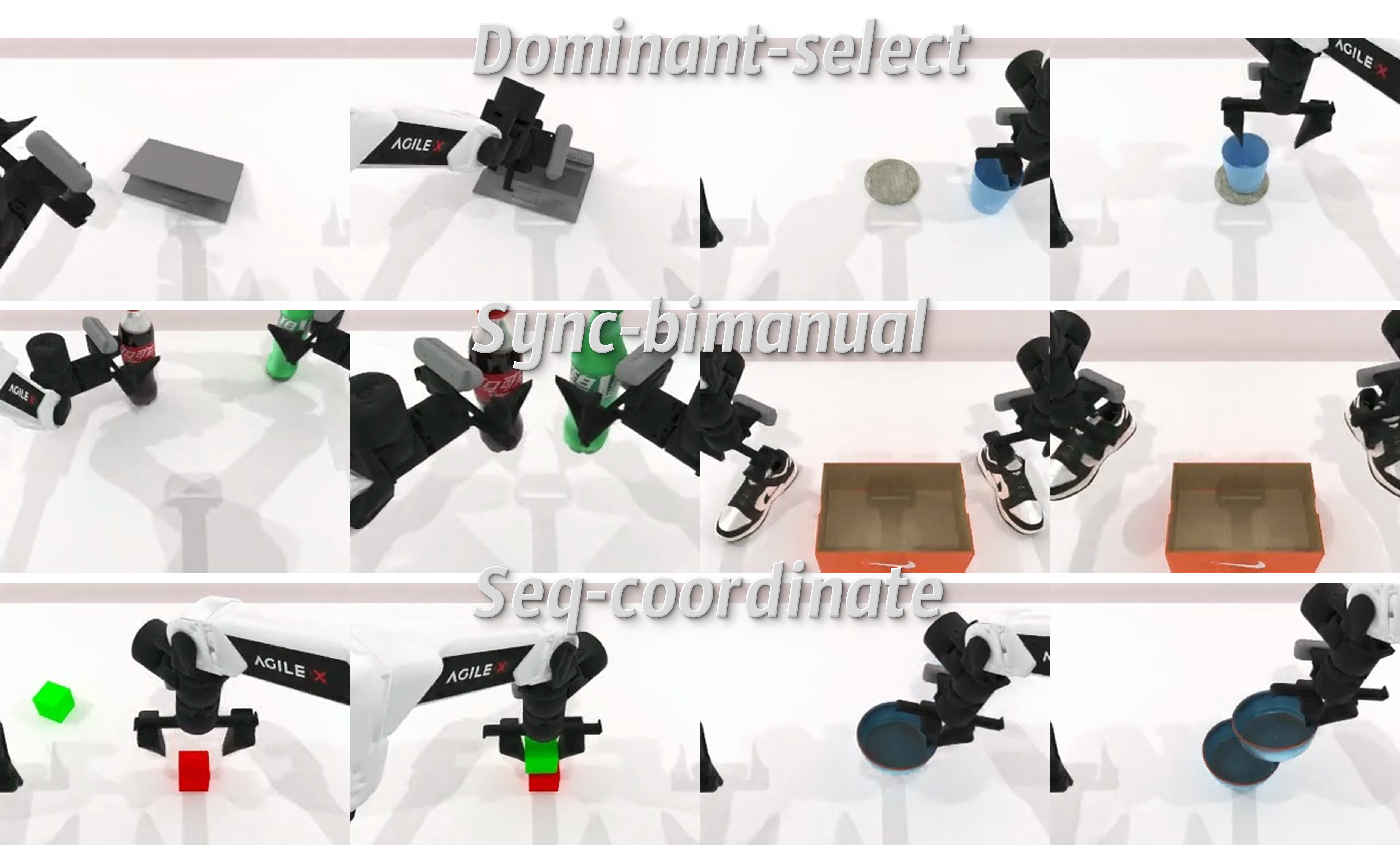}
    \caption{Bimanual tasks in the \textbf{RoboTwin 2.0}~\cite{mu2025robotwin} benchmark. }
    \label{fig:robowin_tasks}
\end{figure}

\subsection{Comparison with State-of-the-Art Methods}
\label{sec:sota_comp}

We first present a comprehensive evaluation in the simulation environment, beginning with a detailed performance breakdown across the three task categories. Subsequently, we provide an analysis of data efficiency.

\noindent\textbf{Benchmark.} We conduct experiments on the \textbf{RoboTwin 2.0}~\cite{mu2025robotwin} benchmark, which provides a diverse set of bimanual manipulation tasks with varying complexity. Following \textit{Xu et al.}~\cite{xu2025imagineact}, we categorize and select the tasks into three types: 16 \textit{Dominant-select} tasks, 8 \textit{Sync-bimanual} tasks, and 8 \textit{Seq-coordinate} tasks to compare policy performance on tasks of varying difficulty, as shown in Fig~\ref{fig:robowin_tasks}.

\noindent\textbf{Implementation Details.} Following \textit{Xu et al.}~\cite{xu2025imagineact}, we train all methods on 100 expert demonstrations in the simulation environment to mitigate distribution gaps. The expert demonstrations are synthesized using 3D generative models and LLMs, starting from \texttt{seed=0}, which facilitates diverse and realistic task variations (approx. 200-1500 steps, 4-30s each). All baselines were reproduced adhering to the default configurations from their original papers or the RoboTwin benchmark. We used varied training schedules (200-600 epochs for 2D methods~\cite{zhao2023learning, chi2023diffusion, xu2025imagineact} and our proposed method; 3000 epochs for 3D methods~\cite{ze20243d, mu2025robotwin}; 150k steps for RDT~\cite{liu_rdt-1b_2025}) with a batch size of 32. For statistical reliability, all methods are trained and evaluated using 3 random seeds. For evaluation, we report the mean and variance of the success rate, calculated over 100 evaluation rollouts for each of 3 random seeds to ensure statistical reliability. All training and evaluations were conducted on 4090 GPUs, except for RDT~\cite{liu_rdt-1b_2025}, which was trained on H20 GPUs; G3Flow~\cite{mu2025robotwin}, which was excluded from some tasks due to its digital twin asset requirements; and \textit{Xu et al.}~\cite{xu2025imagineact}, which encountered NaN errors during training for \textit{Open Microwave} task.

\newcommand{\std}[1]{\text{\scriptsize\textcolor{gray}{\ensuremath{\pm#1}}}}

\begin{table*}[ht]
    \centering
    \caption{\textbf{Comparison on Dominant-select Tasks (16 tasks).} Single-arm manipulation tasks requiring appropriate arm selection. We report the mean and standard deviation of success rates averaged over 3 random seeds. Best score in \textbf{bold}, second-best \underline{underlined}.}
    \footnotesize
    \begin{tabular*}{\textwidth}{@{}r|@{\extracolsep{\fill}}cccccccc@{}}
        \toprule
         & \textbf{\cellcolor{gray!20} Avg.$\uparrow$} & \textup{Beat Block} & \textup{Place} & \textup{Move} & \textup{Open} & \textup{Open} & \textup{Place A2B} & \textup{Place A2B} \\
         \textbf{Method} & \textup{\cellcolor{gray!20}} $(\%)$ & \textup{Hammer} & \textup{Shoe} & \textup{Can Pot} & \textup{Laptop} & \textup{Microwave} & \textup{Left} & \textup{Right}\\
        \midrule
        ACT~\cite{zhao2023learning} & \cellcolor{gray!20} 34.1 & \underline{59.3} \std{6.4} & 14.3 \std{4.0} & 48.0 \std{6.0} & 59.7 \std{2.1} & \underline{83.7} \std{1.5} & 1.3 \std{0.6} & 0.7 \std{0.6} \\
        DP~\cite{chi2023diffusion} & \cellcolor{gray!20} 44.4 & \textbf{{62.0}} \std{4.0} & 43.3 \std{6.7} & 51.0 \std{4.0} & 61.3 \std{3.5} & 53.7 \std{6.5} & 10.0 \std{5.3} & 11.7 \std{5.0} \\
        DP3~\cite{ze20243d} & \cellcolor{gray!20} 61.2 & 59.3 \std{2.3} & 57.7 \std{2.5} & \textbf{{63.7}} \std{1.5} & \underline{77.7} \std{2.1} & 70.3 \std{2.5} & \textbf{{29.3}} \std{2.5} & 23.0 \std{2.6} \\
        G3Flow~\cite{chen2025g3flow} & \cellcolor{gray!20} 60.7 & - & 58.3 \std{1.5} & - & - & - & 25.7 \std{1.5} & \textbf{{25.7}} \std{3.2} \\
        Xu \textit{et al.}~\cite{xu2025imagineact} & \cellcolor{gray!20} 58.4 & 50.0 \std{5.6} & \underline{58.7} \std{9.1} & \underline{58.7} \std{2.1} & 67.7 \std{3.2} & - & 24.3 \std{2.1} & 23.0 \std{2.6} \\
        Ours & \cellcolor{gray!20} 63.2 & 56.3 \std{1.2} & \textbf{{59.7}} \std{1.5} & 55.3 \std{2.1} & \textbf{{88.7}} \std{2.5} & \textbf{{86.0}} \std{1.7} & \underline{26.0} \std{4.4} & \underline{24.7} \std{2.1} \\
        
        \midrule
         \multicolumn{1}{c}{\textup{Turn}} & \textup{Place} & \textup{Place Container} & \textup{Press} & \textup{Place Phone} & \textup{Rotate} & \textup{Place} & \textup{Place Object} & \textup{Shake} \\
         \multicolumn{1}{c}{\textup{Switch}} & \textup{Fan} & \textup{Plate} & \textup{Stapler} & \textup{Stand} & \textup{QRcode} & \textup{Empty Cup} & \textup{Stand} & \textup{Bottle} \\
        \midrule
        \multicolumn{1}{c}{8.7 \std{2.5}} & 2.3 \std{1.2} & 50.7 \std{1.5} & 27.7 \std{1.5} & 43.3 \std{5.8} & 3.0 \std{1.0} & 60.3 \std{2.5} & 1.3 \std{1.2} & 81.7 \std{7.5} \\
        \multicolumn{1}{c}{50.0 \std{4.4}} & 8.0 \std{3.6} & 53.0 \std{6.9} & 28.7 \std{2.5} & 37.7 \std{7.6} & 41.7 \std{3.5} & 62.3 \std{6.7} & \underline{42.0} \std{7.5} & 94.3 \std{1.2} \\
        \multicolumn{1}{c}{\textbf{{55.0}} \std{3.6}} & 46.3 \std{6.0} & 92.3 \std{1.5} & \underline{77.0} \std{2.6} & 52.0 \std{3.6} & \textbf{{76.0}} \std{2.0} & 71.3 \std{2.1} & 34.0 \std{0.0} & \underline{95.0} \std{1.0} \\
        \multicolumn{1}{c}{-} & 51.7 \std{5.7} & 92.3 \std{1.5} & 74.3 \std{2.1} & 45.3 \std{4.0} & 74.3 \std{1.5} & \underline{73.0} \std{7.2} & - & 86.7 \std{2.1} \\
        \multicolumn{1}{c}{48.0 \std{2.6}} & \textbf{{60.0}} \std{2.6} & \underline{93.0} \std{2.0} & 72.3 \std{3.8} & \textbf{{52.7}} \std{2.9} & 68.3 \std{7.2} & 64.3 \std{8.6} & 39.3 \std{4.2} & \textbf{{96.3}} \std{3.5} \\
        \multicolumn{1}{c}{\underline{51.0} \std{1.0}} & \underline{52.0} \std{2.0} & \textbf{{93.3}} \std{0.6} & \textbf{{78.0}} \std{3.0} & \underline{52.3} \std{1.5} & \underline{75.0} \std{4.4} & \textbf{{76.3}} \std{6.0} & \textbf{{42.7}} \std{4.2} & 93.3 \std{2.9} \\
        \bottomrule
    \end{tabular*}
    \label{tab:dominant_select}
\end{table*}
\begin{table*}[ht]
    \centering
    \caption{\textbf{Comparison on Sync-bimanual Tasks (8 tasks).} Bimanual tasks requiring synchronized coordinated operation. We report the mean and standard deviation of success rates averaged over 3 random seeds. Best score in \textbf{bold}, second-best \underline{underlined}.}
    \footnotesize
    \begin{tabular*}{\textwidth}{@{}r|@{\extracolsep{\fill}}ccccccccc@{}}
        \toprule
         & \textbf{\cellcolor{gray!20} Avg.$\uparrow$} & \textup{Pick Diverse} & \textup{Pick Dual} & \textup{Place Bread} & \textup{Place Bread} & \textup{Place Burger} & \textup{Place Cans} & \textup{Place Dual} & \textup{Put Bottles} \\
        \textbf{Method} & \textup{\cellcolor{gray!20}} $(\%)$ & \textup{Bottles} & \textup{Bottles} & \textup{Skillet} & \textup{Basket} & \textup{Fries} & \textup{Plasticbox} & \textup{Shoes} & \textup{Dustbin} \\
        \midrule
        ACT~\cite{zhao2023learning} & \cellcolor{gray!20} 32.4 & 11.7 \std{4.0} & 52.7 \std{3.1} & 12.3 \std{2.5} & 15.7 \std{3.1} & 74.3 \std{2.5} & 29.3 \std{3.5} & 3.0 \std{1.0} & 60.0 \std{4.6} \\
        DP~\cite{chi2023diffusion} & \cellcolor{gray!20} 37.1 & 18.3 \std{3.8} & 52.3 \std{0.6} & 25.0 \std{4.6} & 16.7 \std{3.5} & 76.7 \std{4.2} & 62.7 \std{3.1} & 7.0 \std{4.4} & 38.3 \std{5.1} \\
        DP3~\cite{ze20243d} & \cellcolor{gray!20} 45.1 & \underline{27.7} \std{8.6} & \underline{59.0} \std{3.0} & \underline{36.3} \std{6.7} & \textbf{{23.7}} \std{6.8} & 60.3 \std{3.2} & \underline{66.0} \std{3.6} & 17.7 \std{4.2} & 70.3 \std{2.3} \\
        G3Flow~\cite{chen2025g3flow} & \cellcolor{gray!20} 45.8 & 23.0 \std{3.6} & 58.7 \std{1.2} & 31.0 \std{1.0} & 19.3 \std{4.2} & 73.7 \std{5.0} & \textbf{{67.3}} \std{2.1} & 20.7 \std{4.2} & 73.0 \std{4.6} \\
        RDT~\cite{liu_rdt-1b_2025} & \cellcolor{gray!20} 28.8 & 9.7 \std{1.2} & 44.3 \std{3.8} & 21.7 \std{2.1} & 8.7 \std{3.2} & 65.7 \std{4.0} & 15.7 \std{3.1} & 18.7 \std{3.1} & 45.7 \std{5.9} \\
        Xu \textit{et al.}~\cite{xu2025imagineact} & \cellcolor{gray!20} 47.6 & \textbf{{28.0}} \std{1.0} & 39.7 \std{2.5} & \textbf{{37.7}} \std{3.1} & 15.0 \std{5.3} & \underline{81.7} \std{4.0} & 64.7 \std{1.2} & \underline{40.3} \std{4.2} & \underline{74.0} \std{2.0} \\
        Ours & \cellcolor{gray!20} 51.3 & 24.3 \std{1.2} & \textbf{{69.0}} \std{2.0} & 34.0 \std{6.9} & \underline{20.7} \std{1.2} & \textbf{{82.3}} \std{4.7} & 61.3 \std{6.1} & \textbf{{43.3}} \std{1.2} & \textbf{{75.7}} \std{3.2} \\
        \bottomrule
    \end{tabular*}
    \label{tab:sync_bimanual}
\end{table*}

\noindent\textbf{Quantitative Comparison on Dominant-select Tasks.} These tasks primarily involve single-arm manipulations, which are relatively simple but demand precise 3D spatial reasoning. As shown in Table~\ref{tab:dominant_select}, 2D-based methods (ACT, DP) struggle significantly, while methods explicitly using 3D information, like DP3, achieve a strong and competitive average performance (61.2\%). This gap highlights the necessity of 3D geometric understanding even for these tasks.
Our method achieves the highest average success rate (63.2\%), outperforming all baselines. Notably, our approach surpasses the 3D-native DP3 despite operating only on 2D inputs, demonstrating its ability to effectively infer critical 3D information. This performance gap is most evident in tasks demanding significant spatial inference, such as \textit{Open Laptop} and \textit{Open Microwave}, where our method shows a clear advantage over all 3D-aware baselines.

\noindent\textbf{Quantitative Comparison on Sync-bimanual Tasks.} Sync-bimanual tasks are exceptionally challenging as they require the policy to simultaneously and precisely solve for both arm actions, introducing significant dynamic coupling issues. As shown in Table~\ref{tab:sync_bimanual}, this inherent complexity understandably leads to lower overall performance across the board.
While explicit 3D methods like DP3 and G3Flow generally outperform their 2D counterparts, their performance remains notably limited. DP3's heavy reliance on a fixed, sparse number of sampled points appears to be a critical bottleneck for modeling complex, simultaneous interactions, as seen in its markedly low 17.7\% score on \textit{Place Dual Shoes}.
Our method again secures the top average performance (51.3\%). Its strength in fine-grained bimanual coordination is highlighted by its success in \textit{Place Dual Shoes} (43.3\%), directly overcoming the aforementioned limitation of DP3. This result suggests that our 2D-based predictive paradigm can better capture the complex spatial and temporal relationships required for synchronized control.

\begin{table*}[ht]
    \centering
    \caption{\textbf{Comparison on Seq-coordinate Tasks (8 tasks).} Sequential coordination tasks requiring multi-step bimanual cooperation. We report the mean and standard deviation of success rates averaged over 3 random seeds. Best score in \textbf{bold}, second-best \underline{underlined}.}
    \footnotesize
    \begin{tabular*}{\textwidth}{@{}r|@{\extracolsep{\fill}}ccccccccc@{}}
        \toprule
         & \textbf{\cellcolor{gray!20} Avg.$\uparrow$} & \textup{Handover} & \textup{Hang} & \textup{Place Object} & \textup{Scan} & \textup{Stack} & \textup{Stack} & \textup{Stack} & \textup{Stack} \\
        \textbf{Method} & \textup{\cellcolor{gray!20}} $(\%)$ & \textup{Block} & \textup{Mug} & \textup{Basket} & \textup{Object} & \textup{Blocks Two} & \textup{Blocks Three} & \textup{Bowls Two} & \textup{Bowls Three} \\
        \midrule
        ACT~\cite{zhao2023learning} & \cellcolor{gray!20} 29.4 & 45.3 \std{2.5} & 4.3 \std{0.6} & 6.3 \std{0.6} & 5.7 \std{3.8} & 23.3 \std{3.5} & 0.7 \std{1.2} & 85.3 \std{3.2} & 64.3 \std{3.2} \\
        DP~\cite{chi2023diffusion} & \cellcolor{gray!20} 33.6 & 32.0 \std{1.0} & 15.3 \std{3.2} & 30.0 \std{3.0} & 10.3 \std{2.5} & 17.3 \std{3.8} & 1.3 \std{1.5} & 89.3 \std{1.5} & \underline{73.3} \std{3.1} \\
        DP3~\cite{ze20243d} & \cellcolor{gray!20} 36.0 & 29.7 \std{3.1} & 23.3 \std{7.8} & 32.3 \std{2.5} & 23.3 \std{1.5} & \underline{31.3} \std{3.1} & \textbf{{3.7}} \std{1.5} & 86.7 \std{3.8} & 58.0 \std{6.9} \\
        G3Flow~\cite{chen2025g3flow} & \cellcolor{gray!20} 46.3 & - & \underline{26.7} \std{3.1} & 32.3 \std{2.5} & 22.0 \std{4.6} & - & - & 89.7 \std{3.1} & 61.0 \std{2.6} \\
        RDT~\cite{liu_rdt-1b_2025} & \cellcolor{gray!20} 37.5 & 60.3 \std{2.5} & 25.3 \std{0.6} & 37.0 \std{1.7} & 11.7 \std{1.5} & 24.3 \std{2.1} & 1.7 \std{0.6} & 88.0 \std{2.0} & 51.3 \std{0.6} \\
        Xu \textit{et al.}~\cite{xu2025imagineact} & \cellcolor{gray!20} 45.5 & \underline{64.3} \std{5.8} & 25.0 \std{4.6} & \underline{52.7} \std{5.0} & \underline{28.3} \std{3.1} & 30.7 \std{3.1} & 0.0 \std{0.0} & \underline{90.0} \std{1.0} & 73.3 \std{1.5} \\
        Ours & \cellcolor{gray!20} 50.4 & \textbf{{68.7}} \std{0.6} & \textbf{{40.0}} \std{5.3} & \textbf{{60.3}} \std{3.5} & \textbf{{30.7}} \std{2.1} & \textbf{{32.0}} \std{6.2} & \underline{2.0} \std{1.0} & \textbf{{95.3}} \std{1.5} & \textbf{{74.0}} \std{3.6} \\
        \bottomrule
    \end{tabular*}
    \label{tab:seq_coordinate}
\end{table*} \noindent\textbf{Quantitative Comparison on Seq-coordinate Tasks.} These are the most complex tasks, requiring multi-step bimanual cooperation. Success hinges critically on long-term temporal reasoning and future prediction. The results in Table~\ref{tab:seq_coordinate} confirm this challenge; this demand for foresight is a major failure point for 2D methods like ACT and DP. While 3D-aware methods (DP3, G3Flow) and those with better dependency modeling (RDT) show improved performance, they still struggle with the long-horizon nature of these tasks.
Once again, our method leads with the highest average success rate (50.4\%). Its ability to handle long-horizon planning is best illustrated by the \textit{Hang Mug} task, which rely on precise 3D geometric reasoning. Our method achieves a 40.0\% success rate, significantly outperforming the second 3D-aware methods (G3Flow at 26.7\% and RDT at 25.3\%). This superior performance demonstrates that by focusing on predicting future states, our approach is effective at capturing the spatial-temporal structure and multi-step dependencies required for complex, sequential bimanual actions.

\begin{figure}[t]
    \centering
    \includegraphics[width=\linewidth]{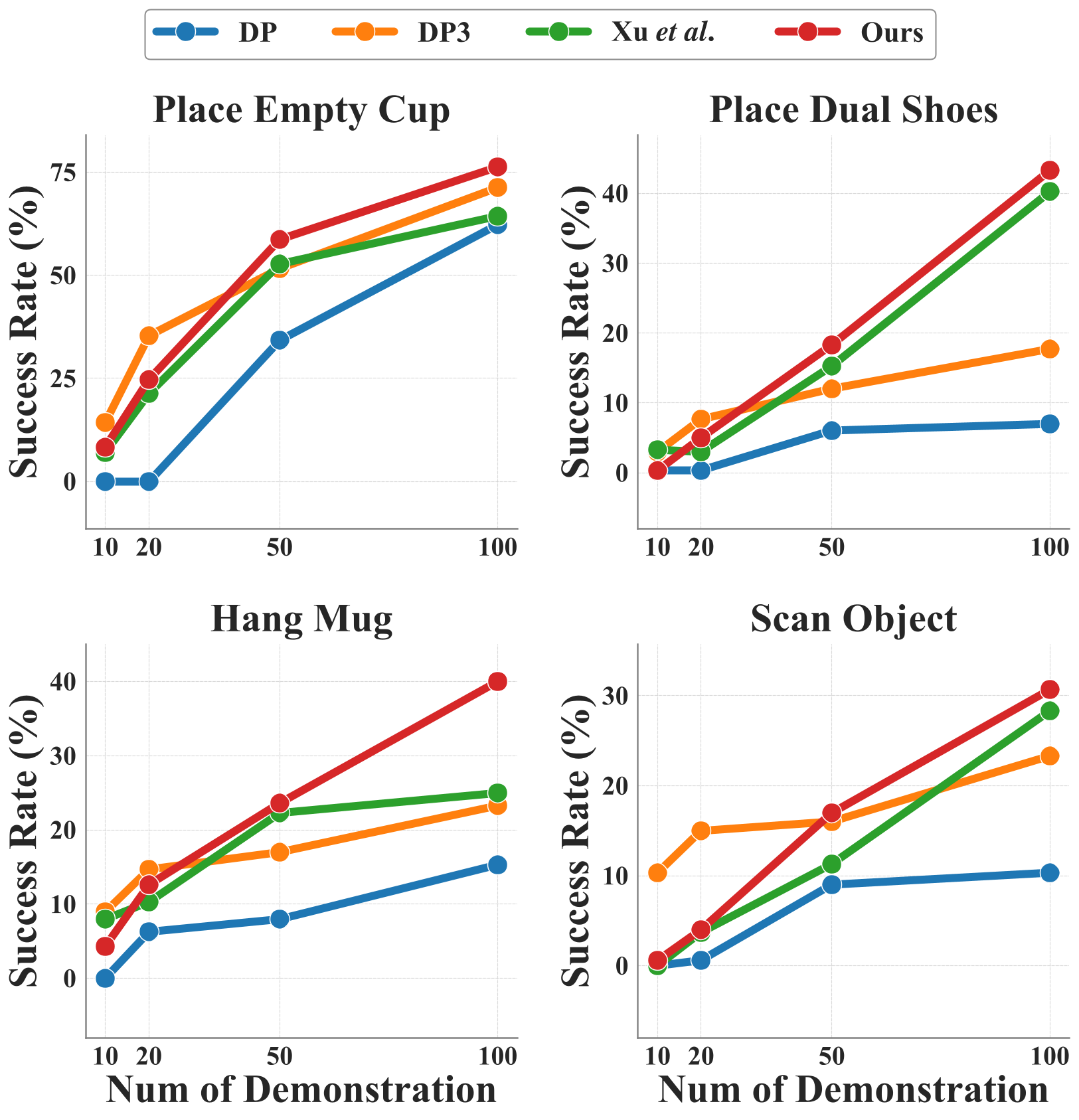}
    \caption{\textbf{Data Efficiency.} Leveraging pre-trained features, our method achieves high data efficiency, outperforming 2D methods in low-data regimes and surpassing the performance of DP3 as more data becomes available.}
    \label{fig:data_efficiency}
\end{figure}

\noindent\textbf{Data Efficiency Analysis.} To evaluate the data efficiency of our method, we conduct experiments on four challenging tasks using a varied number of expert demonstrations: 10, 20, 50, and 100. As illustrated in Figure~\ref{fig:data_efficiency}, our approach demonstrates superior sample efficiency compared to 2D-based methods and better scalability than 3D-based methods.
Specifically, in low-data regimes (10-20 demonstrations), the pre-trained features leveraged by our method provide a significant advantage. In contrast, the 2D baseline DP often fails to learn entirely. For example, on \textit{Scan Object}, DP scores 0.0\% with 10 demos and only 0.6\% with 20. Similarly, on \textit{Place Empty Cup}, its success rate remains at 0.0\% for both 10 and 20 demos. Under these exact same settings, our method already shows a clear learning signal, achieving 24.7\% on \textit{Place Empty Cup} (at 20 demos) and 5.0\% on \textit{Place Dual Shoes} (at 20 demos), demonstrating its stronger foundation for learning from limited data.

\subsection{Ablation Study}
\label{sec:ablation}

We conduct a module-level ablation on four bimanual manipulation tasks: \textit{place empty cup}, \textit{place dual shoes}, \textit{hang mug}, and \textit{scan object}. We run 100 evaluation trials for each of 3 random seeds per task. The metric reported in Tab.~\ref{tab:module_ablation} is the average success rate 4 all tasks.
We systematically disable certain modules to isolate their effects. First, we remove the 2D semantic module to evaluate the importance of semantic guidance. Next, we disable the geometric imagination branch, which predicts the future 3D pointmap, to assess the role of predictive 3D reasoning. Finally, we remove both the 3D geometric module and geometric imagination to test the impact of 3D perception. 
The results show that the full model, with all modules enabled, achieves the highest average success rate of 25.1\%. Removing the 2D semantic module reduces performance slightly to 24.4\%, confirming that semantic features help reasoning about task-relevant objects. Disabling the geometric imagination lowers success to 23.6\%, highlighting that predicting the future 3D pointmap is critical for planning physically plausible bimanual actions. Removing both the 3D geometric module and geometric imagination results in a significant drop to 21.0\%, demonstrating that 3D perception also contributes to effective coordination. These findings indicate that predictive pointmap reasoning is the main driver of performance, while semantic and 3D perception provide complementary benefits.
\begin{table}[t]
\centering
\caption{\textbf{Ablation study.} Impact of removing key components on the average success rate on four tasks.}
\small
\begin{tabular}{ccc c}
\toprule
\makecell{2D Semantic \\ Module} & \makecell{3D Geometric \\ Module} & \makecell{Geometric \\ Imagination} & \makecell{Success \\ Avg.(\%)} \\
\midrule
\checkmark & \checkmark & \checkmark & \textbf{25.1} \\
$\times$   & \checkmark & \checkmark & 24.4 \\
\checkmark & \checkmark & $\times$   & 23.6 \\
\checkmark & $\times$   & $\times$   & 21.0 \\  
\bottomrule
\end{tabular}
\label{tab:module_ablation}
\end{table}

\subsection{Real-World Evaluation}
\label{sec:realworld_comp}

\noindent\textbf{Real-World Experimental Setup.} Our real-world platform consists of the AgileX Cobot Magic bimanual system, equipped with three RealSense D435i cameras. These cameras are strategically mounted to capture synchronized, multi-view RGB images at $480 \times 640$ resolution, providing comprehensive visual observation of the workspace. We evaluate our method on four challenging bimanual manipulation tasks from the RoboTwin 2.0 benchmark: \textit{Scan Object}, \textit{Hang Mug}, \textit{Stack Three Bowls}, and \textit{Place Dual Shoes}.  For each task, we collect a dataset of 50 expert demonstrations via teleoperation. To evaluate performance, we execute 20 trials for each task and report the average success rate. During this real-world deployment, all policy inference is performed on a workstation equipped with an RTX 4090 GPU. 


\noindent\textbf{Experimental Results.} As shown in Table \ref{tab:realworld_comp}, our method outperforms all baselines on all four real-world bimanual manipulation tasks. Our method achieves an average success rate of 40\%, significantly surpassing all baseline methods, including ACT (23.8\%), DP (25\%), and \textit{Xu et al.} (32.5\%).
The advantage of our method is particularly pronounced on the most challenging tasks. For instance, on the \textit{Hanging Mug} task, both ACT and DP methods failed completely, whereas our method achieved a 20\% success rate. Similarly, for the \textit{Place Dual Shoes}, our method significantly outperforms all comparison methods, particularly ACT, which failed the task entirely. Even on the comparatively simpler \textit{Place Empty Cup} task, our method still outperforms the next-best method. These real-world experimental results strongly demonstrate the robustness and superior performance of our method in handling complex, high-precision bimanual tasks.
\begin{table}[t]
    \centering
    \caption{\textbf{Real-world experiments.} Success rates (\%) of different methods on four bimanual manipulation tasks.}
    \small
    \begin{tabular}{lcccc}
    \toprule
    Task & ACT & DP & Xu \textit{et al.} & Ours \\
    \midrule
    Place Empty Cup    & 70 & 70 & 75 & 80 \\
    Place Dual Shoes   & 0 & 10 & 15 & 20 \\
    Hanging Mug        & 0 & 0 & 5 & 20\\
    Scan Object        & 25 & 20 & 35 & 40 \\
    \midrule
    \textbf{Avg. Success (\%) $\uparrow$} & 23.8 & 25 & 32.5 & \textbf{40} \\
    \bottomrule
    \end{tabular}
    \label{tab:realworld_comp}
\end{table}

\begin{figure}[t]
    \centering
    \includegraphics[width=\linewidth]{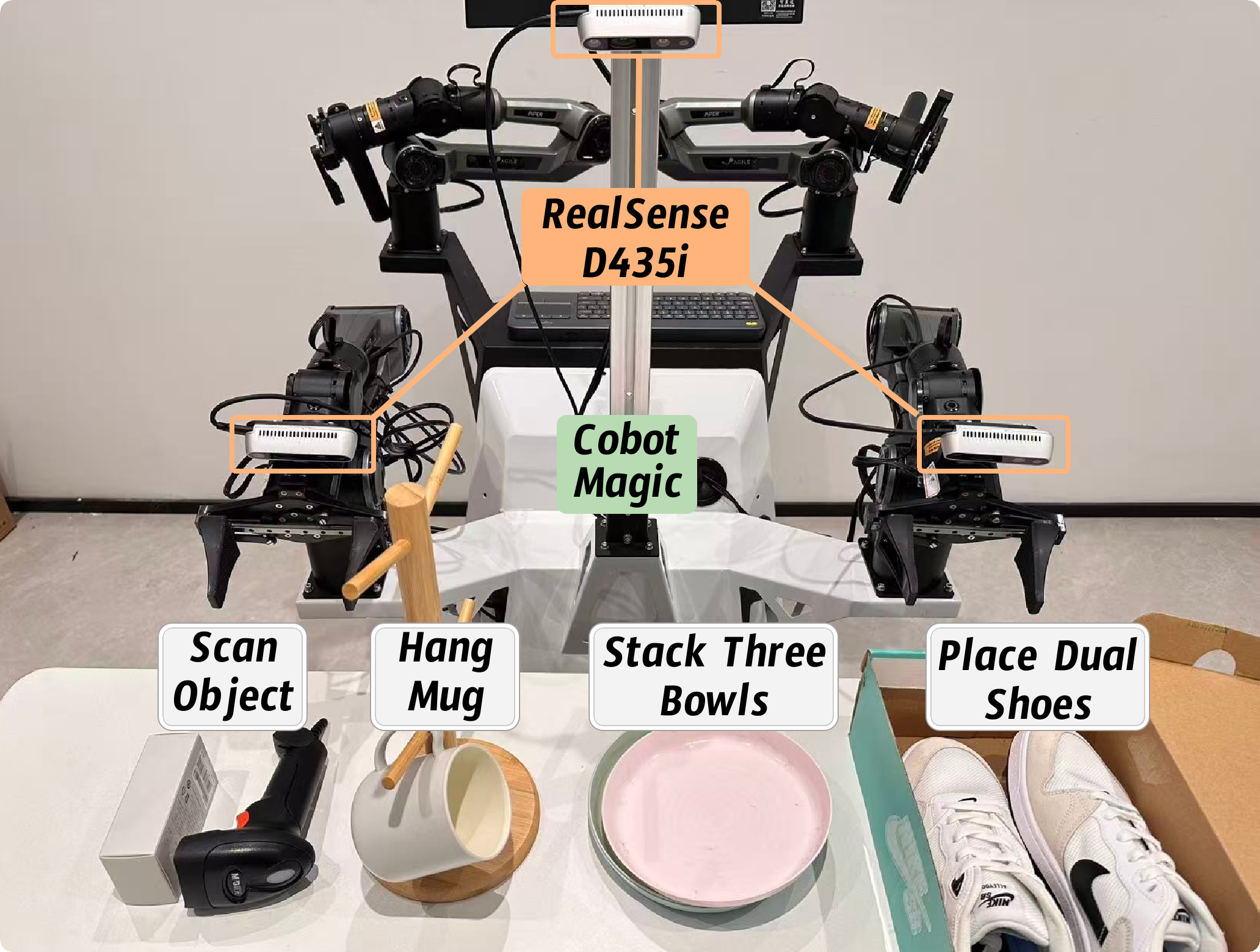}
    \caption{\textbf{Real-World Setting.} Our real-world platform featuring the AgileX Cobot Magic bimanual system, equipped with three RealSense D435i cameras to evaluate four challenging tasks.}
    \label{fig:real_setting}
\end{figure}
\label{sec:real_exp}

\section{Conclusion}
\label{sec:conclusion}

We presented a bimanual manipulation framework leveraging a pre-trained 3D geometric foundation model to achieve RGB-only, 3D-aware control. By fusing 2D semantic cues, 3D geometric latents, and robot proprioception, and jointly predicting future action chunks with a future 3D pointmap latent, our policy acquires strong spatial understanding and predictive capability. Extensive experiments in simulation and real world demonstrate our approach outperforms 2D-based and point-cloud-based baselines in manipulation success, coordination quality, and 3D spatial prediction accuracy. Ablation studies further confirm the importance of geometric perception and predictive 3D reasoning paradigm.

\noindent\textbf{Limitation and future work.} Our model's single-step predictive horizon and lack of persistent 3D memory constrain long-term state accumulation and reasoning. Future work will focus on extending predictions to multi-step 3D trajectories, improving temporal coherence, and broadening generalization to unseen tasks and objects.

{\small
\bibliographystyle{ieeenat_fullname}
\bibliography{references}
}

\end{document}